\documentclass[10pt,twocolumn,letterpaper]{article}

\usepackage{iccv}
\usepackage{times}
\usepackage{epsfig}
\usepackage{graphicx}
\usepackage{amsmath}
\usepackage{amssymb}
\usepackage{caption}

\usepackage{multirow}

\usepackage[pagebackref=true,breaklinks=true,letterpaper=true,colorlinks,bookmarks=false]{hyperref}

 \iccvfinalcopy 

\def\iccvPaperID{xxxx} 

\ificcvfinal\fi

\begin{document}

\title{Learning Garment DensePose for Robust Warping in Virtual Try-On}

\author{
Aiyu Cui$^{1,2}$\thanks{Work done during an internship at Meta.}
\and
Sen He$^{1}$
\and
Tao Xiang$^{1}$
\and
Antoine Toisoul$^{1}$
\and \newline
{ $^1$ Meta AI} ~ 
{ $^2$ University of Illinois at Urbana-Champaign, USA} ~
\\
{\tt\small aiyucui2@illinois.edu \quad \{senhe, txiang, atoisoul\}@meta.com}
}

\maketitle

\begin{abstract}

  Virtual try-on, i.e making people virtually try new garments, is an active research area in computer vision with great commercial applications. Current virtual try-on methods usually work in a two-stage pipeline. First, the garment image is warped on the person's pose using a flow estimation network. Then in the second stage, the warped garment is fused with the person image to render a new try-on image. Unfortunately, such methods are heavily dependent on the quality of the garment warping which often fails when dealing with hard poses (e.g., a person lifting or crossing arms). In this work, we propose a robust warping method for virtual try-on based on a learned garment DensePose which has a direct correspondence with the person's DensePose. Due to the lack of annotated data, we show how to leverage an off-the-shelf person DensePose model and a pretrained flow model to learn the garment DensePose in a weakly supervised manner. The garment DensePose allows a robust warping to any person's pose without any additional computation. Our method achieves the state-of-the-art equivalent on virtual try-on benchmarks and shows warping robustness on in-the-wild person images with hard poses, making it more suited for real-world virtual try-on applications.
\end{abstract}

\section{Introduction}

E-commerce sales are increasing every year estimated to reach a stunning 6.3 trillion US dollars in 2023. Although the growth of e-commerce has accelerated following the pandemic, customers are often uncertain about the products they are buying leading to a high amount of returns, especially when it comes to fashion items. Virtual try-on attempts to solve this issue that can be defined as the task of making people virtually try new garments. In other words, given two inputs, a person image and a new garment image, the try-on task consists in generating a photorealistic image of the person wearing the new garment. It is a popular area of research driven by its high commercial potential for online shopping platforms. 

State-of-the-art virtual try-on pipelines~\cite{ge2021parserfree, he2022styleflow, yang2020acgpn} treat the virtual try-on task as two subtasks: warping the new garment image on the person's pose and then fusing the warped garment with the person to create the final try-on image. For the garment warping stage, previous works~\cite{chopra2021zflow, he2022styleflow,yang2020acgpn} train a network conditioned on both the garment and person images in order to learn a geometric transformation (a flow field) between the garment and person images. However such flow prediction networks often result in inaccurate warpings for person images with hard poses which are underrepresented in the training set or with complex limb gestures. This limitation is mainly due to the design choice of the warping network. First, the fact that it is conditioned on the person image, makes the flow field estimation highly dependent on the person's pose. In addition, due to the intrinsic limitation of convolutional neural networks, it is difficult to predict significantly different offsets in the flow field for connected regions in the garment, e.g., upper sleeve and lower sleeve. As a result, current state-of-the-art garment warping networks are still not robust enough to generalize to in-the-wild images.

In this paper, we overcome this issue by stepping away from directly learning a pose dependent geometric transformation between the garment and person images. Instead, we leverage the DensePose information of the human body, which already contains an accurate dense correspondence with every point on the body surface, to get an accurate warping whose quality is not affected by the various poses encountered in real-life applications.
 
More specifically, our method learns a DensePose for garments from garment images in a weakly supervised manner. From the predicted garment DensePose, the garment image can be warped onto any new person pose using the correspondence between the garment and person DensePose through UV space. Unlike previous methods, since the garment DensePose is directly predicted from the garment image, the diversity of human poses does not affect the prediction's accuracy, leading to a more robust warping. We learn our proposed model on a simple dataset (VITON-HD~\cite{choi2021vitonhd}), but apply it to hard cases to show its robustness.

\textbf{In summary we make the following contributions:}

\begin{itemize}
   \item We introduce a garment DensePose for garment-only images that directly correspond with a person's DensePose, allowing virtual try-on in more complex and realistic cases (Section 3).
  
  \item The garment DensePose is directly learned in a weakly supervised manner, i.e., without requiring manually annotated garment DensePose data which is both not available for garments and tedious to acquire.
  
  \item We demonstrate that our method achieves equivalent performance with state-of-the-art virtual try-on methods on VITON-HD benchmarks and validate the superior robustness of the proposed warping method on human in hard poses (Section 4). 

\end{itemize}

\section{Related Work}

\paragraph{Image-based virtual try-on} usually follows the \textit{garment warping} with \textit{image fusion} paradigm. The garment image is first warped to align with person image and then fused with the person image using a U-Net~\cite{ronneberger2015u} to generate the final try-on image. Based on the garment warping methods employed in different works, current image-based virtual try-on can be categorized into either Thin Plate Spline (TPS)~\cite{duchon1977splines} or appearance flow~\cite{zhou2016view} based methods.

TPS based methods \cite{han2018viton, wang2018toward, minar2020cp,yang2020acgpn, ge2021disentangled, yang2022rt-vton} predict a non-rigid deformation grid to warp the garment. \cite{han2018viton} used shape context matching to estimate the TPS transformation. \cite{wang2018toward} exploited the correlation between the feature map of the person representation and the garment feature map to estimate the TPS transformation. \cite{ge2021disentangled} used a predicted target garment mask to guide the prediction of the TPS transformation parameters. \cite{yang2022rt-vton} proposed a semi-rigid deformation to combine the flexibility of TPS with the rigidity of affine transformation.

Appearance flow based methods~\cite{han2019clothflow, ge2021parserfree, li2021ovnet, chopra2021zflow, cui2021dressing, he2022styleflow, dong2022dressing-in-the-wild} predict the offset for each pixel in the garment image to warp it. Compared to TPS based methods, appearance flow based methods have more degree of freedom for garment warping and thus have been adopted by most recent state-of-the-art try-on works. \cite{han2019clothflow} applied the hierarchy in the feature map to predict the appearance flow. \cite{ge2021parserfree} added a feature correlation module proposed in optical flow estimation~\cite{dosovitskiy2015flownet, ilg2017flownet} for appearance flow estimation. \cite{he2022styleflow} proposed style-based appearance flow estimation. 

In contrast to TPS and appearance flow based methods, our work proposes a garment DensePose which uses the correspondence between the garment and person images in the fixed IUV space~\cite{guler2018densepose}. It avoids the direct dependence on the person image for garment warping as in TPS and appearance flow based methods and is more robust and applicable to in-the-wild and realistic virtual try-on settings. Note that~\cite{albahar2021pose-with-style, dong2022dressing-in-the-wild} also applied a similar idea in a different virtual try-on setting: garment swap\footnote{Garment swap, in which the garment is already on the person's body. Therefore, predicting the garment DensePose is not required as it can be easily and directly extracted from its mask predicted using a human parser.}. Compared to theirs, our setting is more difficult as we need to predict the ghost mannequin garment DensePose in an unsupervised manner due to the lack of annotation in the current benchmarks.

\vspace{-5mm}
\paragraph{3D virtual try-on} 3D virtual try-on fits a garment image onto a 3D human body. It can provide a better try-on experience compared to image based virtual try-on. Traditional methods~\cite{bhatnagar2019multi, mir2020learning} used SMPL~\cite{loper2015smpl} to model a 3D garment template, which however, requires 3D human body scans and thus cannot be easily scaled up. Recent work~\cite{zhao2021m3d} used a 2.5D representation (2D try-on with an additional depth map for 3D rendering) for 3D virtual try-on. However, it still requires to employ a TPS based garment warping.

\vspace{-5mm}
\paragraph{Human body surface representations}
Human body surface models parameterize the surface of the human body. DensePose~\cite{guler2018densepose} is the pioneer work that represents the human body in 24 parts based on the SMPL model~\cite{loper2015smpl}. In DensePose, each location on the human body surface is assigned with a body part label and a 2D UV location. \cite{neverova2020continuous} (CSE) extended DensePose into a continuous space and also applied it to represent the body of other common animals. \cite{ianina2022bodymap} proposed a transformer based approach to estimate high resolution CSE for human body. \cite{xie2022temporaluv} proposed TemporalUV which alleviated the problem of DensePose for loose garments. Our work exploits the IUV space defined in DensePose as an intermediate space to find the correspondence between the garment and person images. The proposed method is less affected by hard person poses when warping the garment and is, therefore, better suited for in-the-wild and realistic image-based virtual try-on settings.
\section{Methodology}

We formally define the virtual try-on task in this work. Given a garment product image $g \in \mathbb{R}^{3\times H \times W}$ (garment only without a model) and a person image $p \in \mathbb{R}^{3\times H \times W}$, a virtual try-on model generates a try-on image $t \in \mathbb{R}^{3\times H \times W}$ which has the given person in $p$ wearing the garment $g$. 

Our virtual try-on pipeline is composed of three steps. First, we extract the garment foreground mask for garment images, human parsing for person images and person DensePose from person images which have face masked out by off-the-shelf models \cite{Neverova2019DensePoseConfidences, li2020schp} (so no facial DensePose is detected or used). 
Next, we predict garment DensePose with our proposed method. The predicted garment DensePose, together with person DensePose, is used to warp the garment image. Finally, a blending model fuses the warped garment with a masked person image to generate the final try-on image.

\vspace{-3mm}
\subsection{Preliminary: DensePose}

DensePose~\cite{guler2018densepose} is a parametrization of the human body surface that divides the human body into $24$ body parts each having its own UV space. More specifically, given an image with a human subject, for each pixel on the person, DensePose provides a $I$ label indicating which body part the pixel belongs to and a $(u,v)$ coordinate label for the UV correspondence. Note that identical body parts for two different persons will always have the same UV coordinates in the DensePose parametrization, regardless of the pose, body shape, and view angles. To avoid ambiguity, we denote by $I_{p}$, $U_{p}$, and $V_{p}$ the person DensePose body part segmentation, U coordinates and V coordinates respectively, predicted by a off-the-shelf person DensePose model $\mathbf{M_{p}}$ \cite{guler2018densepose}:
\begin{equation}
    I_p, U_p, V_p = \mathbf{M_{p}}(p),
    \label{eq:iuv_pred}
\end{equation}

\subsection{Weakly Supervised Garment DensePose Prediction}

The main limitation of previous SOTA warping methods \cite{han2019clothflow, he2022styleflow} is that they need to predict a flow field for garment warping conditioned on the person image. This causes them to easily fail for hard person poses. To this end, in this work, we introduce garment Densepose for garment warping. The proposed garment DensePose only relies on the garment image itself and can directly be used to warp the garment image by finding its correspondence with the person DensePose in the fixed IUV space.

One of the main challenges to predicting garment DensePose is the lack of annotation in the current VITON benchmarks~\cite{han2018viton,dong2019mpv}. The only weakly supervised label we can obtain is the masked person DensePose ($\overline{I_{p}}, \overline{U_{p}}, \overline{V_{p}}$), i.e., the area corresponding to garment region in the predicted person DensePose:
\begin{equation}
    \overline{I_{p}} = I_{p} \otimes m_p^g,  \ \overline{U_{p}} = U_{p} \otimes m_p^g, \ \overline{V_{p}} = V_{p} \otimes m_p^g,
\end{equation}
where $m_p^g$ is the garment mask predicted from the off-the-shelf human parser~\cite{li2020schp} on the person image and $\otimes$ is element-wise multiplication.

It is thus straightforward to directly use the masked garment ($p_{m} = p \otimes m_p^g$) from the person image as an input to train a garment DensePose prediction model with masked person DensePose as a weakly supervised label. However, due to the distribution shift between masked garments in person images and the real garment images, the learned model does not perform well when applied to real garment images (see ablation study in the experiment section).

To make the best use of the available supervision from the off-the-shelf person DensePose prediction model, we propose a novel indirect garment DensePose training method. Concretely, given a garment-only image $g$, we use a garment DensePose model, $\mathbf{M_{g}}$, to predict the DensePose for garment image:
\begin{equation}
    I_g, U_g, V_g = \mathbf{M_{g}}(g),
    \label{eq:giuv_pred}
\end{equation}
where $ I_g, U_g \ \text{and} \ V_g \in \mathbb{R}^{25 \times H \times W}$ are the garment part segmentation and the prediction of $u,v$ coordinates in UV texture space for each class respectively. 

As the predicted garment DensePose is not spatially aligned with the weakly supervised label, we still cannot train $M_{g}$ using $\overline{I_{p}}, \overline{U_{p}}, \overline{V_{p}}$. To this end, we propose to use the pretrained flow model from Style-Flow~\cite{he2022styleflow} $\mathbf{F}$ to predict a flow field $f = \mathbf{F}(g, p)$ between the garment image $g$ and person image $p$. With the predicted flow field $f$, the predicted garment DensePose can be warped to align with the garment region in the predicted person DensePose:
\begin{equation}
    \hat{I_g} = \omega(I_g,f),\hat{U_g} = \omega(U_g,f), \hat{V_g} = \omega(V_g,f),
\end{equation}
where $\omega(\cdot, \cdot)$ is a bi-linear sampling based warping. Note that $\mathbf{F}$ is only used during the training of $\mathbf{M_{g}}$ as warping the predicted garment DensePose is not needed during inference.

To train the garment DensePose model $\mathbf{M_{g}}$, we first compute the IUV loss ($\mathcal{L}_{IUV}$) with the predicted garment DensePose:
\begin{equation}
    \begin{aligned}
        & \mathcal{L}_{IUV} (\hat{I_g},\hat{U_g}, \hat{V_g},U_g,V_g) \\
        & \ = \mathcal{L}_{cls}(\hat{I_{g}}, \overline{I_{p}}) + \mathcal{L}_{l1}(\hat{U_{g}}, \overline{U_{p}}) +  \mathcal{L}_{l1}(\hat{V_{g}}, \overline{V_{p}})\\
        & \ + \mathcal{L}_{tv}(U_g)  +\mathcal{L}_{tv}(V_g),
    \end{aligned}
\end{equation}
where $\mathcal{L}_{cls}$ is the cross entropy loss, $\mathcal{L}_{l1}$ is the L1 distance and $\mathcal{L}_{tv}$ is the total variation loss \cite{ianina2022bodymap} to preserve the smoothness of the predicted UV coordinates.

\subsection{Warping via Garment DensePose}

In principle, given a garment DensePose and a person DensePose, we can spatially warp the garment on the person's body using the correspondence provided by the UV space. However, in practice, due to the DensePose sparsity and low resolution of the texture map, this naive approach for warping gives poor results. In this subsection, we will dig into these two problems and introduce solutions to improve the warping. 

\subsubsection{Inpainting and Nearest Neighbor Sampling for UV Sparsity Issue}
The first challenge to tackle is the sparsity of the UV coordinates in the DensePose space. More specifically, the garment and person DensePoses rarely have a perfect overlap of UV coordinates to build a one-to-one mapping between the two. In particular, this creates empty values in the warped result whenever there is a small shift of coordinates between the two DensePoses, i.e. between the known and query UV coordinates. As a result, a naive warping leads to large missing areas, as not every query UV coordinate (in the person DensePose) has an existing value in the UV texture extracted from the garment DensePose.

Pose-with-style \cite{albahar2021pose-with-style} proposed a solution to the DensePose sparsity issue in the case of pose transfer (warping a source person image to a new pose) by inpainting the texture to span the full UV space so that any query UV coordinate has a corresponding value. Unfortunately, in our virtual try-on setting, this inpainting is not directly applicable. Indeed, inpainting the texture to span the entire UV space also inpaints unwanted regions and yields to inaccurate results, such as turning a short sleeve top into a long-sleeved one. It is therefore key to preserve the garment shape as accurately as possible when inpainting the UV space and restrict the inpainted region to the original shape of the garment. 

We, therefore, propose a new solution to alleviate the sparsity issue. Rather than inpainting the full UV space, we predict a mask $m_{q}$ to restrict the inpainting to the masked region. This mask is obtained in two steps. First, we get a coarse mask $m'_q$ on the person in 2D image space by warping the mask of the original garment image via the naive DensePose alignment. This gives us a sparsely warped mask, which roughly describes the shape of the warped garment but contains a noisy boundary and holes. Then, we train a neural network $\mathbf{R}$ to refine this warped garment mask as: 
\begin{equation}
    m_q = \mathbf{R}(m'_q).
\end{equation}
where $R$ has the holes filled and the noisy garment boundary smoothed. We train $\mathbf{R}$ with a binary cross entropy loss $\mathcal{L}_{bce}(m_q, m_p^g)$ with the preprocessed garment mask on person image $m_p^g$ as the supervision. 

This refined warped garment mask $m_{q}$ can be projected into UV space as a query mask. More specifically, for any UV pixel inside the query mask that does not have a valid source value, we choose the nearest valid source value. Finally, the full garment can be warped using this mask guided-inpainted UV map, which allows us to ensure that there is no hole in the warped garment.

\subsubsection{Coordinate Warping for Resolution Issue}
We further find that the resolution of the UV map highly affects the warping quality. With a low-resolution UV texture map, the UV texture is less sparse but the warped garment is more blurry due to the downsampling operation, and conversely. In other words, there is a trade-off between the warped garment resolution and the sparsity of the UV texture map. Even using the restricted inpainting introduced above to mitigate the UV sparsity issue, we would always be in favor of a less sparse texture map, due to the fact that the inpainting accuracy can never be guaranteed. 

Following prior work~\cite{albahar2021pose-with-style, dong2022dressing-in-the-wild}, instead of directly warping the RGB pixel value to UV space, we warp the coordinate grid (i.e., the $(x,y)$ positions) of the source garment image and then use the warped grid to sample the RGB pixels directly from the source garment image. This way, we can avoid information loss due to the low resolution of the UV texture map which gives a more stable warping result as the upsampling and downsampling operations do not hurt the coordinate grid as much as it does on raw texture. We denote the warped garment as $g_{warp}$. 

\subsection{Try-On Image Generation}
After the garment is correctly warped to the given person's pose, the next step is to blend the warped garment $g_{warp}$ with the person together. We follow the parser-based fusion pipeline of~\cite{ge2021parserfree, he2022styleflow} and train a generator $\mathbf{G}$ which takes as input the processed person image $p'$ with the upper body masked out and filled with mean skin color, 
the warped garment image $g_{warp}$ and the DensePose mask $m_p^{arm}$ for the region of the arms and hands that do not overlap with the warped garment. $\mathbf{G}$ then predicts both a blending mask $\alpha$ and a coarse try-on $\hat{t}$ image:

\begin{equation}
    \hat{t}, \alpha = \mathbf{G}(p', g_{warp}, m_p^{arm})
\end{equation}

The final result $t$ is a blend between the warped garment and the coarse try-on result as $t = (1 - \alpha)\hat{t}  + \alpha g_{warp}$. This blending step allows bringing high frequency details back in the garment region of the final result.

Following prior work, we train $\mathbf{G}$ with a L1 Loss $\mathcal{L}_{l1}$, a Perceptual Loss \cite{johnson2016perceptual-loss} $\mathcal{L}_{perc}$ and a Style Loss~\cite{gatys2016style-loss-1, gatys2015style-loss-2} $\mathcal{L}_{sty}$ between the generated image and the ground truth for both $t'$ and $t$. Note that, during training, the person in $p$ is wearing the same garment as in $g$. In addition, we add a blending mask regularization loss as used in~\cite{he2022styleflow} but change the L1 regularization to L2 as we empirically found that it leads to better image quality.

Moreover, following \cite{yang2020acgpn, yang2022rt-vton}, we find train the generator $\mathbf{G}$ with an inpainting task will further improve the result, so we randomly mask out the processed person input $p'$ during the training with free-form masks computed by the algorithm from Yu et al. \cite{yu2019freeform}.

\section{Experiments}
\subsection{Implementation Details}
\paragraph{Dataset}
We experiment with our method on the VITON-HD ~\cite{choi2021vitonhd} and MPV benchmarks~\cite{dong2019mpv}. VITON-HD has $11,647$ training pairs and $2,032$ test pairs of person and garment-only images at $1024 \times 768$ resolution, which we resize to $256 \times 192$ for all our experiments. 

For MPV, as the garment input only have their frontal view displayed, following prior work~\cite{ge2021parserfree}, we remove all back view person images. After processing, the MPV dataset includes $35,687$ person images and $13,524$ clothes images at resolution $256 \times 192$, including $4,175$ image pairs for the test set. 
To validate our model's warping robustness for hard poses, we further select a subset of the test split of MPV, denoted MPV-hard, that contains hard poses. In this subset, we pick full-body person images which are very different from VITON-HD’s 3/4 body length images with simple poses. Our selected MPV-hard test set contains a total of 1,670 pairs. Compared to VITON-HD, which mostly contains simple pose person images with a 3/4 body length frontal setup, the person images in MPV-hard contain a full body with various arm poses.

When we compute person DensePoses in the preprocessing steps, we first mask out the faces based on the human parsing before sending it to the DensePose predictor for all person images in all datasets, so no facial DensePose is detected or used.

\paragraph{Network Implementation and Training Details} 
Our model is implemented in PyTorch and trained with an A100 GPU on AWS.
For preprocessing, we use the SCHP model~\cite{li2020schp} to get human parsing in the LIP label format and we detect DensePose with~\cite{Neverova2019DensePoseConfidences} for human images. 
We implement the garment DensePose prediction model $\mathbf{M}_g$ with a U-Net architecture~\cite{ronneberger2015u} and train it for $20$ epochs with a learning rate of $1e-4$, a batch size of 32 and a $0.1$ learning rate decay every 5 epochs. For the fusion model $G$ and the query mask refinement model $R$, we use the same residual U-Net architecture as in StyleFlow~\cite{he2022styleflow}. These are trained with a batch size of $16$ and a learning rate of $1e-4$ for 80 epochs with a learning rate decay of $0.1$ from epoch 40.  

\paragraph{Metrics} To measure the performance of the virtual try-on task, following the convention, we use Structural Similarity Index Measure (SSIM)~\cite{wang2004ssim} to evaluate the generation accuracy and the Fr\'echet Inception Distance (FID)~\cite{heusel2017fid} for the generation realism. 
To evaluate warping robustness, i.e, warping quality on person images in hard poses, we propose Normalized Masked SSIM (NM-SSIM), which is the SSIM of the garment region in the generated try-on image normalized by the percentage of the union area of the warped garment and the ground truth garment segmentation. We also compute the mean IoU between areas of the warped garment and the garment segmentation in person images that pair with the input garment image. 

\subsection{Comparisons}
\paragraph{Benchmark Test} We compare our method with the following state-of-the-art virtual try-on methods on VITON-HD benchmark \cite{choi2021vitonhd}: CP-VTON \cite{wang2018cp-vton}, ACGPN\cite{yang2020acgpn}, HD-VITON~\cite{choi2021vitonhd}, HR-VITON~\cite{lee2022hrviton}, Cloth-Flow\cite{han2019clothflow}, and StyleFlow~\cite{he2022styleflow}. 
Note that most compared methods are parser-based (including our method) except for StyleFlow~\cite{he2022styleflow} which is a parser-free method that requires a complex training and distillation strategy. For a fair comparison, we re-trained StyleFlow~\cite{he2022styleflow} in a parser-based manner using their official code. 

Tab.~\ref{tab:vitonhd_sota} shows that our method outperforms most methods and gets equal performance to StyleFlow\cite{he2022styleflow} and ClothFlow \cite{han2019clothflow}. 
\begin{table}[h!]
\centering
\caption{Quantitative results of different models on VITON-HD \cite{choi2021vitonhd} benchmark in resolution $256\times 192$. $^*$ indicates that StyleFlow is re-trained in a parser-based manner for a fair comparison. PR is the short for ``Preference Rate''.}
\begin{tabular}{ |c|c|c|c| } 
\hline
Methods & SSIM $\uparrow$       & FID $\downarrow$ & PR(theirs/ours) \\ 
 \hline
 CP-VTON \cite{wang2018cp-vton}                 & 0.739          & 30.11 & N/A \\ 
 HD-VITON \cite{choi2021vitonhd}         & 0.811          & 16.36  & N/A \\
 ACGPN \cite{yang2020acgpn}                     & 0.833          & 11.33  & N/A \\  
 HR-VITON \cite{lee2022hrviton}         & 0.864          & {9.38}  & 29.9\% / \textbf{70.1\%} \\
 Cloth-Flow \cite{han2019clothflow}             & 0.857          & 9.48  & \textbf{56.0\%} / 44.0\% \\ 
$^*$StyleFlow \cite{he2022styleflow}            & 0.857          & 9.45   & \textbf{51.6\%} / 48.3\% \\

 \hline
 Ours                                          & \textbf{0.867}     & \textbf{9.19}  & - \\
 \hline
 
\end{tabular}
\vspace{-5mm}
\label{tab:vitonhd_sota}
\end{table}

\paragraph{Robustness Test} 
To compare the warping robustness, we train flow-based state-of-the-art methods Cloth-Flow~\cite{han2019clothflow}, StyleFlow~\cite{he2022styleflow} and HR-VITON \cite{lee2022hrviton} on VITON-HD dataset \cite{choi2021vitonhd} and directly test them on MPV-hard datasets\footnote{This is closer to the real-world setting where we train the model on a relatively clean dataset and then apply it to in-the-wild person images.}. In this case, all trained models learn the warping from the simple frontal pose humans in the VITON-HD dataset and apply the learned models to the challenging MPV-hard dataset which has more diverse person inputs in full body with various arm gestures.

From the quantitative results shown in Tab.~\ref{tab:mpv_hard}, the proposed method has significantly better performances than all the compared methods in the out-of-distribution cases and complex pose cases. 
These results prove that unlike the previous warping methods which overfit the training pose distributions by learning flow estimators' parameters from the paired garment and person, the proposed method can effectively avoid this training bias by separately extracting the DensePoses for both garment and person and computing the flow mathematically from these. Therefore, our method provides a more robust warping, that can be learned from simple data and applied to difficult cases in real-world scenarios. 

\begin{table}
\centering
\caption{Quantitative results of warping robustness of different methods. MPV-hard indicates that results are only tested on the selected hard pose person images. All methods are trained VITON-HD\cite{choi2021vitonhd} with simple poses. PR is the short for ``Preference Rate''.}
\vspace{-3mm}
\begin{tabular}{ |c|c|c|c| } 
\hline
\multirow{2}{*}{Methods} &  \multicolumn{3}{c|}{MPV-hard} \\ 
\cline{2-4}
& NM-SSIM & mIoU & PR (theirs/ours) \\
 \hline
 HR-VITON~\cite{lee2022hrviton}& 0.204 &0.511 & 14.5\% / \textbf{85.9\%} \\
 Cloth-Flow~\cite{han2019clothflow}& 0.290 & 0.760 & 29.9\% / \textbf{70.1\%} \\
 $^*$StyleFlow~\cite{he2022styleflow}& 0.282 & 0.741 & 26.5\% / \textbf{73.5\%} \\
 \hline
 Ours &\textbf{0.337} &\textbf{0.827} & - \\
 \hline
\end{tabular}
\vspace{-7mm}
\label{tab:mpv_hard}
\end{table}

\paragraph{User Study}
 Aside from automatic metrics, we also run a user study to manually evaluate our method. We randomly select 500 test pairs as the candidate pools for VITON-HD and MPV-Hard respectively. For each compared method in HR-VITON~\cite{lee2022hrviton}, Clothflow~\cite{han2019clothflow}, Style-Flow~\cite{he2022styleflow}, we show users the input and output from two unlabeled models (one is ours) and ask their preferences. The results are collected from 10 volunteers who have work experience in generative models. 
 
  The user study results in Tab.~\ref{tab:vitonhd_sota} are consistent with our quantitative results. Our method is slightly less favorable in the user study compared with StyleFlow and ClothFlow, because the flow interference causes nonsmooth textures as identified in Sec. 4.4. However, this discrepancy is not significant and still shows that we have similar performance to the SOTA on the VITON-HD benchmark along with the SSIM and FID.

\subsection{Ablation Study} In this part, we validate our training objective of Equation~\ref{eq:l_aux}, our warping method of section 3.3, and the training objectives of the generator $\mathbf{G}$. Results are shown in Table~\ref{tab:giuv_table}.

\vspace{-0.4cm}
\paragraph{A. Learning garment DensePose from the cropped garment.} We propose to learn the garment DensePose prediction model using indirect supervision from garments in the garment-only image by a pre-trained flow model. 
To validate the necessity of this design,  we compare with the baseline that learns garment DensePose from cropped garments $p_g$ of person images $p$ as $I_{p_g}, U_{p_g},V_{p_g} = \mathbf{M}_g(p_g)$, so there exists direct supervision:
\begin{equation}
    \begin{aligned}
        \mathcal{L}_{base} & = \mathcal{L}_{cls}(I_{p_{g}}, \overline{I_{p}}) + \mathcal{L}_{l1}(U_{p_{g}}, \overline{U_{p}}) +  \mathcal{L}_{l1}(V_{p_{g}}, \overline{V_{p}})\\
        & \ + \mathcal{L}_{tv}(U_{p_{g}})  +\mathcal{L}_{tv}(V_{p_{g}}).
    \end{aligned}
    \label{eq:l_aux}
\end{equation}
Note, we set the background of the cropped garment to white pixels during the training to better mimic the test case (garment-only images). 

The ablation study shows that the naive baseline results in an inaccurate warping at test time, because of the domain gap between the cropped garment and the garment-only images, which leads to a worse result in Tab. \ref{tab:giuv_table}.

\vspace{-0.4cm}
\paragraph{B. Mask-guided UV inpainting, nearest neighbor sampling and coordinate warping} 
In the second ablation study, we justify the proposed warping methods of Sec 3.3. 
1) We remove the inpainting in UV space guided by the predicted warping mask from the full pipeline.
2) We also remove the grid sampling, in which case, the RGB pixels from the garment image are warped to the UV space and then to the person's pose.
The guided inpainting is necessary for filling holes created in DensePose warping and the grid sampling is essential to preserve the texture sharpness.

\vspace{-0.4cm}
\paragraph{C. Fusion} Though we inherit the same fusion architecture and losses from~\cite{he2022styleflow}, we add a style loss~\cite{gatys2015neural} and add an inpainting mechanism during the training. In Table~\ref{tab:giuv_table} we ablate the two changes separately and show that the two changes are indeed necessary.
\begin{table}
\centering

\caption{\textbf{Ablation Studies} on VITON-HD dataset \cite{choi2021vitonhd} at resolution $256 \times 192$.}
\vspace{-3mm}
\begin{tabular}{ |l|l|ll| } 
\hline
\multicolumn{2}{|c|}{ }                      & SSIM $\uparrow$ & FID $\downarrow$ \\   
\hline
\hline

A & learn from cropped garment  &  0.860  &  10.01  \\ 

\hline
 \multirow{2}{*}{B} & w/o guided UV inpainting          & 0.857  & 11.45 \\
                        & w/o grid warping        &  0.863 & 12.15   \\ 
 \hline
 
 \multirow{2}{*}{C} & w/o style loss          &  0.865 & 10.85 \\
                  & w/o inpainting               & 0.860 & 10.34      \\ 
 \hline 
\multicolumn{2}{|c|}{\bf full model}           &  \textbf{0.867} & \textbf{9.19}  \\
 \hline
\end{tabular}
\vspace{-2mm}
\label{tab:giuv_table}
\end{table}

\subsection{Failure Cases} Although our model achieves the equivalent performance as state-of-the-art on benchmark test and show significant robustness over all previous methods, it has scope for improvements. In particular, we identify two main limitations in our experiments. 
First, as our warping method highly depends on the quality of the off-the-shelf person DensePose prediction model~\cite{guler2018densepose}, any inaccurate person DensePose detection, especially in the garment region, leads to inaccurate warping and artifacts in the final generated try-on images. 
Second, unlike previous methods which predict a single flow field that directly samples the garment from the source image to the target person, the proposed method requires a two-stage sampling by first warping to UV space and then warping from the UV space to target RGB space based on the corresponding DensePoses. This two-stage sampling amplifies non-smooth flow artifacts and distorts complex patterns in garments.
Consequently, in the user study, results are less favorable, especially for garments with stripes and checkered patterns.

\section{Conclusion}
In this work, we have proposed a novel garment DensePose that can be employed to warp garments in a virtual try-on task. Our method is trained using weakly supervised labels without any manual annotation and achieves state-of-the-art equivalent results. Overall, the method is more robust to the real-word person images with complex poses and is therefore suited for real-world try-on applications.

\newpage
{\small
\bibliographystyle{ieee_fullname}
\bibliography{egbib}
}

\newpage


\end{document}